\documentclass{article}

\usepackage{arxiv}

\usepackage[utf8]{inputenc} 
\usepackage[T1]{fontenc}    
\usepackage{hyperref}       
\usepackage{url}            
\usepackage{booktabs}       
\usepackage{amsfonts}       
\usepackage{amsmath}        
\usepackage{nicefrac}       
\usepackage{microtype}      
\usepackage{graphicx}
\usepackage{float}
\graphicspath{{./images/}}

\title{Evolutionary Algorithms for Computing Nash Equilibria in Dynamic Games}

\author{
  Alireza Rezaee \\
  School of Intelligent Systems\\
  College of Interdisciplinary Science and Technology\\
  University of Tehran\\
  Tehran, Iran\\
  \texttt{arrezaee@ut.ac.ir} \\
}

\begin{document}
\maketitle

\begin{abstract}
Dynamic nonzero sum games are widely used to model multi agent decision making in control, economics, and related fields. Classical methods for computing Nash equilibria, especially in linear quadratic settings, rely on strong structural assumptions and become impractical for nonlinear dynamics, many players, or long horizons, where multiple local equilibria may exist. We show through examples that such methods can fail to reach the true global Nash equilibrium even in relatively small games. To address this, we propose two population based evolutionary algorithms for general dynamic games with linear or nonlinear dynamics and arbitrary objective functions: a co evolutionary genetic algorithm and a hybrid genetic algorithm particle swarm optimization scheme. Both approaches search directly over joint strategy spaces without restrictive assumptions and are less prone to getting trapped in local Nash equilibria, providing more reliable approximations to global Nash solutions.
\end{abstract}


\section{Introduction to Game Theory}
\label{sec:introduction}

In non-cooperative games where players act simultaneously (i.e., without a hierarchical order or prior knowledge of other players' moves at each stage), the goal is to find an equilibrium point at which all players simultaneously optimize their value functions (costs or payoffs). In many such games, at each decision stage, the actions of all players influence not only their own objective functions, but also those of other players. This is in contrast with standard single-objective optimization problems, where only a single control input (decision variable) is optimized.

At each stage of a dynamic game, all players select their strategies. Each player's action affects its own payoff and also the payoffs of the other players. In order to make good decisions and optimize its payoff, each player must choose a strategy that is optimal given the strategies of all the other players.

Thus, unlike standard optimization problems---where each objective is treated separately and optimized with respect to its own variables---in dynamic games the control is exerted over a class of functions that jointly affect all players' objectives and must be optimized simultaneously. This leads naturally to the concept of a Nash equilibrium \cite{BasarOlsder1999}.

Computing Nash equilibria in dynamic games is challenging. Classical approaches rely on dynamic programming, Riccati equations in linear--quadratic (LQ) settings, or first-order necessary conditions, and they become increasingly intractable as the dimension of the state space, the number of players, and the time horizon grow \cite{Ozyildirim2000}. Moreover, these methods are tightly coupled to specific model structures (e.g., linear dynamics and quadratic costs), whereas many modern applications in networked systems, intelligent transportation, and cyber-physical systems exhibit nonlinear, hybrid, or data-driven dynamics \cite{BarolliAINA2024,BarolliBWCCA2019,BarolliWAINA2019}.

In parallel, evolutionary computation and swarm intelligence have been successfully applied to a wide range of optimization problems in signal processing \cite{Rezaee2010FIR}, communications and antennas \cite{MohamadzadeRezaee2017Antenna,Taghvaee2014Metamaterial}, control systems \cite{RezaeeGolpayegani2012,Rezaee2017PID,Rezaee2017MPC,Rezaee2017Penetrometer}, cloud and distributed computing \cite{Rezaee2014FuzzyCloud,Gavagsaz2018LoadBalancing}, and smart grids or energy systems \cite{Barzamini2012}. These successes motivate the use of evolutionary algorithms for dynamic games, where the goal is to search over joint strategy spaces rather than single-objective design variables.

\subsection{Definition of an $N$-Player Dynamic Game}

\textbf{Definition 1.}
Consider an $N$-player discrete-time dynamic game with a finite and known time horizon $K \in \mathbb{N}$. Such a game is characterized by:
\begin{itemize}
    \item A set of players: $\mathcal{N} = \{1,2,\dots,N\}$.
    \item A finite set of stages: $\mathcal{K} = \{0,1,\dots,K\}$, where $K$ is the maximum number of moves (stages).
    \item The system dynamics:
    \begin{equation}
      x_{k+1} = f_k\big(x_k, u_{1,k}, \dots, u_{N,k}\big), \quad k=0,\dots,K-1,
    \end{equation}
    where $x_k$ is the state at stage $k$, and $u_{i,k}$ is the control (action) of player $i$ at stage $k$.
    \item For each player $i \in \mathcal{N}$, a control sequence (strategy) over the horizon:
    \begin{equation}
      u_i = \{u_{i,0}, u_{i,1}, \dots, u_{i,K-1}\}.
    \end{equation}
    \item An information structure for each player, specifying what information is available at each stage. Depending on the information pattern, different solution concepts and equilibrium points may arise. A global Nash equilibrium typically assumes complete information about the system and the objective functions of all players.
    \item For each player $i$, a cost (or payoff) function over the horizon:
    \begin{equation}
      J_i(u_1,\dots,u_N) = \sum_{k=0}^{K-1} g_{i,k}\big(x_k, u_{1,k},\dots, u_{N,k}\big) + h_i(x_K),
    \end{equation}
    where $g_{i,k}$ and $h_i$ encode the running and terminal costs (or rewards).
\end{itemize}

We denote by $u_{-i}$ the collection of strategies of all players except player $i$.

\subsection{Nash Equilibrium}

\textbf{Definition 2.}  
A strategy profile $u^{\star} = (u^{\star}_1,\dots,u^{\star}_N)$ is a (feedback) Nash equilibrium of the game if and only if, for every player $i \in \mathcal{N}$,
\begin{equation}
  J_i\big(u^{\star}_i, u^{\star}_{-i}\big) 
  \leq J_i\big(u_i, u^{\star}_{-i}\big),
  \quad \forall\, u_i.
\end{equation}
In words, given that all other players use $u^{\star}_{-i}$, player $i$ cannot improve its cost (or payoff) by unilaterally deviating from $u^{\star}_i$.

Search techniques for Nash equilibria are often derived from this definition. A Nash equilibrium satisfies the necessary optimality conditions for each player conditional on the strategies of other players. In general, however, for dynamic games with nonlinear dynamics, nonlinear objective functions, and many players and stages, classical methods (e.g., solving first-order necessary optimality conditions) become difficult to implement and may converge only to local equilibria \cite{Pavlidis2005,Wiegand2001,Denzau}. Moreover, many classical methods rely on restrictive assumptions and approximations (e.g., linearization), and one cannot always claim that they find the true global Nash equilibrium.

These limitations have motivated the development of evolutionary and learning-based methods for computing Nash equilibria in more complex games, including coevolutionary strategies \cite{Wiegand2001,SipperGP} and hybrid metaheuristics \cite{SonBaldick2004,MouserDunn2005}.

\section{Evolutionary Approaches to Dynamic Games}

Evolutionary game theory and computational intelligence-based methods provide flexible tools for computing equilibria in dynamic games without strong structural assumptions. Genetic algorithms (GAs) and particle swarm optimization (PSO) are two well-established evolutionary methods that can be adapted to search for Nash equilibria \cite{GenCheng1997,DianatiGAIntro,WhitleyGATutorial,ParsopoulosVrahatis2004}.

A key advantage of these approaches is that they can be applied to a broad class of problems, regardless of:
\begin{itemize}
    \item Linear or nonlinear dynamics,
    \item Number of players or stages,
    \item Shape of the objective functions.
\end{itemize}
In contrast, many mathematical solution methods are effective only for particular game structures. For example, dynamic programming and Riccati-equation-based methods are typically limited to linear--quadratic dynamic games \cite{Ozyildirim2000,BasarOlsder1999}.

Similar ideas have already proven effective in a wide range of pattern recognition and computer vision tasks, such as face detection and recognition \cite{Hajati2006FaceLocalization,Pakazad2006FaceDetection,Hajati2010PoseInvariant,Hajati2017DynamicTexture}, surface geodesic pattern analysis \cite{Hajati2017Surface}, offline signature verification and palmprint recognition \cite{AbdoliHajati2014,Shojaiee2014Palmprint}, and dynamic texture modeling \cite{CremersACCV2014}. In biomedical and healthcare applications, evolutionary and deep models have been used for disease prediction \cite{Mahajan2024}, diabetes therapy initiation \cite{Fiorini2019}, cardiac arrhythmia detection \cite{Sadeghi2024ECG}, fungal image analysis \cite{Sopo2021DeFungi}, COVID-19 screening and related software tools \cite{Sadeghi2024COVID,Tavakolian2022FastCOVID,Tavakolian2022SoftwareImpacts}, and robust machine learning toolkits \cite{Wang2022SoftwareImpacts}. These diverse applications highlight the robustness of evolutionary and learning-based optimization when dealing with high-dimensional, noisy, or non-convex search spaces.

In this paper, we present two population-based evolutionary methods:
\begin{enumerate}
    \item A co-evolutionary genetic algorithm (GA) for Nash equilibrium search.
    \item A hybrid PSO-based method with local search refinement.
\end{enumerate}
We illustrate the performance of these methods with numerical examples and compare them with classical mathematical techniques in the spirit of previous comparative studies between GA and PSO \cite{MouserDunn2005,Hassan2005}.

\section{Genetic Algorithm for Nash Equilibrium Search}
\label{sec:ga}

In this section we present an evolutionary algorithm based on a co-evolutionary genetic algorithm for searching global Nash equilibria in dynamic games, extending previous coevolutionary GA frameworks \cite{SonBaldick2004,Wiegand2001} to multi-stage dynamic settings.

\subsection{Chromosome Encoding}

Each member of the population is a chromosome representing a candidate strategy profile across all players and all stages. If the game has $N$ players and $K$ stages, and each $u_{i,k}$ is a scalar, then the total number of variables is $N K$. A chromosome can be represented as:
\begin{equation}
  \mathbf{c} = \big(u_{1,0},\dots,u_{1,K-1}, u_{2,0},\dots,u_{N,K-1}\big).
\end{equation}

Unlike standard binary encoding, we use a decimal (base-10) encoding scheme. The sign of each variable is encoded using the first digit (e.g., digits 0--4 represent a positive sign, and 5--9 represent a negative sign), while the remaining digits encode the magnitude with user-specified precision. The number of digits and the position of the decimal point are set by the user depending on the desired accuracy, consistent with typical GA engineering design frameworks \cite{GenCheng1997,DianatiGAIntro}.

\subsection{Co-Evolutionary Structure}

The GA is implemented in a co-evolutionary framework: for each player, we maintain a subpopulation of chromosomes focusing on that player's strategy variables, while treating the current best strategies of the other players as fixed \cite{SonBaldick2004,Wiegand2001}.

The main steps are:
\begin{enumerate}
    \item Initialize a random population of chromosomes (or use user-provided initial guesses).
    \item For player $i = 1,\dots,N$:
    \begin{enumerate}
        \item Retrieve the best strategies of all players from the previous iteration.
        \item Apply GA operators (selection, crossover, mutation) only to the variables corresponding to player $i$'s strategies.
        \item Evaluate the fitness for each chromosome using player $i$'s cost function $J_i$.
        \item Update the best strategy of player $i$ and share it with the other players.
    \end{enumerate}
    \item Repeat Step 2 until a stopping criterion is met.
\end{enumerate}

\subsection{Fitness Evaluation}

The fitness of each chromosome is computed based on the players' cost functions. For a maximization GA, we define a positive fitness function, e.g.,
\begin{equation}
  F_i(\mathbf{c}) = C - J_i(\mathbf{c}),
\end{equation}
where $C$ is a sufficiently large constant such that $F_i(\mathbf{c}) > 0$ for all chromosomes. For minimization problems, this transformation allows us to use the standard roulette-wheel selection \cite{WhitleyGATutorial}.

\subsection{Selection, Crossover, and Mutation}

\paragraph{Selection.}  
We use roulette-wheel selection proportional to fitness to select parent chromosomes. If elitism is enabled, the best chromosome is copied directly to the next generation, while the rest of the population is filled by roulette-wheel selection \cite{WhitleyGATutorial}.

\paragraph{Crossover.}  
With crossover probability $P_c$, two parents are selected and a one-point crossover is performed: a random cut point is chosen, and the segments are exchanged between the parents to produce two offspring. Very high $P_c$ (close to 1) may lead to excessive disruption of good solutions, while very low $P_c$ may slow convergence \cite{GenCheng1997,DianatiGAIntro}.

\paragraph{Mutation.}  
Each gene in each chromosome is mutated with a small probability $P_m$, typically in the range
\[
  0.01 \leq P_m \leq 0.2,
\]
depending on the problem. Too large $P_m$ can cause excessive randomness and prevent convergence, while too small $P_m$ may lead to premature convergence to local equilibria \cite{WhitleyGATutorial}.

A mutation randomly changes the encoded digit(s) of a variable within the allowed range. Since the algorithm is co-evolutionary, mutation only affects the variables corresponding to the current player's strategies, and keeps other players' best strategies intact.

\subsection{Stopping Criteria}

The algorithm stops when either:
\begin{itemize}
    \item The maximum number of generations $G_{\max}$ is reached, or
    \item The change in the best fitness over the last $M$ generations is smaller than a threshold $\varepsilon$.
\end{itemize}
Choosing $\varepsilon$ too small may lead to excessive computation and potential convergence to local optima. In practice, a combination of a maximum number of generations and a reasonable fitness tolerance is used.

\subsection{Co-Evolutionary GA Algorithm}

\noindent\textbf{Algorithm: Co-evolutionary GA for Dynamic Games}
\begin{enumerate}
    \item Initialize a random population of chromosomes.
    \item \textbf{For} player $i = 1$ to $N$:
    \begin{enumerate}
        \item Obtain the current best strategies for all players.
        \item Apply GA operators (selection, crossover, mutation) to player $i$'s variables.
        \item Evaluate player $i$'s fitness for all chromosomes.
        \item Update and broadcast player $i$'s best strategy.
    \end{enumerate}
    \item \textbf{Repeat} Step 2 until a stopping criterion is satisfied.
    \item Output the set of best strategies for all players in the final generation as the approximate Nash equilibrium.
\end{enumerate}

\section{Particle Swarm Optimization and Hybrid PSO}
\label{sec:pso}

Particle swarm optimization (PSO) is another population-based evolutionary algorithm inspired by the collective behavior of social organisms such as birds, fish, and insects. PSO is particularly attractive due to its conceptual simplicity and relatively fast convergence \cite{KennedyEberhart1995,Fleischer2003,Hassan2005}.

\subsection{Neighborhood Structures}

Several neighborhood structures can be used in PSO:
\begin{itemize}
    \item \textbf{Global (stellar) neighborhood:} Each particle is connected to all others and tends to follow the best particle in the entire swarm (global best, $g_{\text{best}}$).
    \item \textbf{Local (ring) neighborhood:} Each particle is connected only to a small set of neighbors (e.g., a ring structure), and follows the best particle in its neighborhood (local best).
    \item \textbf{Star or hub-and-spoke neighborhood:} One central particle is connected to all others and broadcasts improvements. This is less commonly used in practice.
\end{itemize}

In this paper, we use the global neighborhood structure, which usually yields faster convergence and dense communication among particles, similar to the global-best PSO variants studied in \cite{ParsopoulosVrahatis2004,LaskariPSOminimax}.

\subsection{Standard PSO Update Equations}

Let the position and velocity of particle $p$ at iteration $t$ be denoted by $\mathbf{x}_p(t)$ and $\mathbf{v}_p(t)$, respectively. Let $\mathbf{pbest}_p$ be the best position found so far by particle $p$, and $\mathbf{gbest}$ be the best position found by any particle in the swarm. The velocity and position updates are:
\begin{align}
  \mathbf{v}_p(t+1) &= \omega\, \mathbf{v}_p(t)
  + c_1 r_1 \big(\mathbf{pbest}_p - \mathbf{x}_p(t)\big)
  + c_2 r_2 \big(\mathbf{gbest} - \mathbf{x}_p(t)\big),
  \\
  \mathbf{x}_p(t+1) &= \mathbf{x}_p(t) + \mathbf{v}_p(t+1),
\end{align}
where:
\begin{itemize}
    \item $\omega$ is the inertia weight,
    \item $c_1$ and $c_2$ are acceleration coefficients,
    \item $r_1$ and $r_2$ are random numbers uniformly distributed in $[0,1]$.
\end{itemize}

Velocity clamping is often used to restrict the maximum speed:
\begin{equation}
  -v_{\max} \leq v_{p,j}(t) \leq v_{\max},
\end{equation}
for each component $j$ of the velocity vector. Large $v_{\max}$ increases global exploration but may overshoot the optimum; small $v_{\max}$ restricts motion to a small region and may slow convergence.

The inertia weight $\omega$ is typically decreased over time, e.g.,
\begin{equation}
  \omega(t) = \omega_{\max} - 
  \frac{\omega_{\max} - \omega_{\min}}{T_{\max}}\, t,
\end{equation}
to encourage exploration during early iterations and exploitation near convergence \cite{Hassan2005}.

\subsection{Parameter Settings}

The PSO parameters must be tuned carefully:
\begin{itemize}
    \item Acceleration coefficients $c_1$ and $c_2$ are often chosen such that $c_1 + c_2 < 4$ to guarantee convergence.
    \item The inertia weight $\omega$ is typically in the range $[0.4, 0.9]$.
    \item The maximum velocity $v_{\max}$ should be set with respect to the scale of the decision variables.
\end{itemize}

PSO does not require selection, crossover, or mutation operators. This simplifies implementation and often yields higher convergence speed compared to GA \cite{MouserDunn2005,Hassan2005}. Furthermore, since PSO does not rely on roulette-wheel selection, the objective function need not be strictly positive.

\subsection{Hybrid PSO with Local Search}

To enhance performance, we combine PSO with a local search method, resulting in a hybrid PSO algorithm. After updating the positions of the particles using the standard PSO update, we apply a local search (e.g., \texttt{fminsearch} in MATLAB) to refine the positions for each player, analogous to hybrid evolutionary schemes used in engineering design and test generation \cite{Rezaee2008GeneticSymbiosis,Rezaee2010FIR,Rezaee2017PID}.

For player $i$, suppose the new PSO-generated position is $\mathbf{x}_p^{(i)}$. We then apply a local optimization method for a fixed number of iterations (denoted by \texttt{HybridIter}) starting from $\mathbf{x}_p^{(i)}$ to obtain an improved position $\tilde{\mathbf{x}}_p^{(i)}$. This refined position is then used to update $\mathbf{pbest}_p$ and $\mathbf{gbest}$.

To reduce the risk of being trapped in local minima, we also introduce occasional mutation: if $\mathbf{gbest}$ does not improve over several iterations, we randomly perturb a subset of particles, similar in spirit to mutation-based diversity injection in evolutionary and swarm-based algorithms \cite{ParsopoulosVrahatis2004,SipperGP}.

\subsection{Hybrid PSO Algorithm}

\noindent\textbf{Algorithm: Hybrid PSO for Dynamic Games}
\begin{enumerate}
    \item Initialize a random swarm of particles, where each particle encodes the strategies of all players.
    \item \textbf{For} player $i = 1$ to $N$:
    \begin{enumerate}
        \item Retrieve the current best strategies for all players.
        \item Update the velocities and positions of the particles corresponding to player $i$.
        \item Apply local search (e.g., \texttt{fminsearch}) to refine the particle positions for player $i$ for \texttt{HybridIter} steps.
        \item Evaluate the fitness of each particle for player $i$ and update $\mathbf{pbest}$ and $\mathbf{gbest}$.
    \end{enumerate}
    \item If $\mathbf{gbest}$ does not improve for a predefined number of iterations, perform random mutation on a subset of particles.
    \item Repeat Step 2 until a stopping criterion is satisfied.
    \item Output the best strategies for all players as the approximate Nash equilibrium.
\end{enumerate}

\section{Numerical Examples}
\label{sec:examples}

In this section we present numerical examples to illustrate the performance of the proposed GA and PSO-based algorithms for computing Nash equilibria.

\subsection{Example 1: Three-Player, Three-Stage Linear-Quadratic Game}

Consider a three-player dynamic game with three stages ($K = 3$), linear dynamics, and quadratic cost functions. Each player's cost depends on its own control inputs and the system states across stages. The dynamic programming solution with complete information and symmetric players yields a reference Nash equilibrium, which we denote by $u^{\text{DP}}$ \cite{BasarOlsder1999,Ozyildirim2000}.

We then apply the co-evolutionary GA and the PSO-based method to search for Nash equilibria numerically, following the approach of evolutionary Nash search in \cite{Pavlidis2005,SonBaldick2004}.

\subsubsection{GA Results}

For the GA, we experimented with different population sizes. Increasing the population size improved the convergence rate (fewer generations required) but increased the computational cost. For population sizes larger than about 40, there was little further improvement, in line with typical GA behavior reported in \cite{GenCheng1997,DianatiGAIntro}.

Using an error tolerance of $10^{-3}$ on the fitness, the GA converged to a solution $\hat{u}^{\text{GA}}$ that closely matches the dynamic programming solution. Due to the use of elitism, the best member is never lost, leading to a smooth convergence of the fitness function.

\begin{figure}[H]
  \centering
  \includegraphics[width=0.4\linewidth]{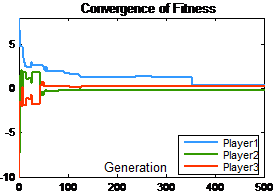}%
  \caption{Convergence of the fitness function for the three-player, three-stage LQ game using the co-evolutionary GA (complete information). The smooth decrease is due to elitism, which preserves the best chromosome in each generation.}
  \label{fig:ga_fitness_triple}
\end{figure}

\subsubsection{PSO Results}

For the PSO algorithm, we tuned the inertia weight, acceleration coefficients, and velocity limits to achieve convergence in approximately 1300 iterations (averaged over 10 runs). The final solution $\hat{u}^{\text{PSO}}$ also closely matched the dynamic programming solution, with variation across runs smaller than $0.01$, consistent with the robustness of global-best PSO reported in \cite{ParsopoulosVrahatis2004,MouserDunn2005}.

\begin{figure}[H]
  \centering
  \includegraphics[width=0.4\linewidth]{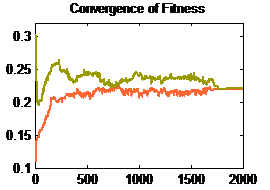}%
  \caption{Convergence of the PSO algorithm for the three-player LQ game: fitness versus iteration.}
  \label{fig:pso_fitness_triple}
\end{figure}

\begin{figure}[H]
  \centering
  \includegraphics[width=0.4\linewidth]{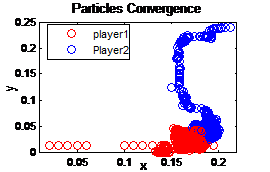}%
  \caption{Trajectories of particle positions (strategies) for the three players in the search space as PSO converges to the Nash equilibrium in the three-player LQ game.}
  \label{fig:pso_strategies_triple}
\end{figure}

The added mutation in the PSO algorithm improved fine-tuning around the optimum and helped avoid local equilibria. In this example, the classical dynamic programming solution fails to satisfy the Nash conditions in some cases (e.g., changing one player's strategy while keeping others fixed decreases that player's cost), while the evolutionary solutions satisfied the Nash conditions more robustly.

\subsection{Example 2: Kydland Two-Player Dynamic Game with Non-Quadratic Objectives}

We next consider the well-known Kydland two-player dynamic game with non-quadratic objective functions and feedback information structure. Each player observes the state and chooses a feedback strategy at each stage \cite{BasarOlsder1999}.

\subsubsection{GA with Feedback Information}

We first solve the problem under feedback information using the GA. The strategies are parameterized as linear feedback laws, and the GA is applied to optimize the parameters. With $10{,}000$ generations, the GA converges to a solution that satisfies the Nash conditions within an error tolerance of approximately $0.05$ in the objective functions across repeated runs.

\begin{figure}[H]
  \centering
  \includegraphics[width=0.4\linewidth]{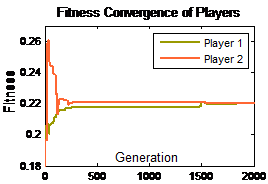}%
  \caption{Convergence of the GA fitness function for the two-player Kydland dynamic game with complete information.}
  \label{fig:ga_fitness_kydland}
\end{figure}

\begin{figure}[H]
  \centering
  \includegraphics[width=0.4\linewidth]{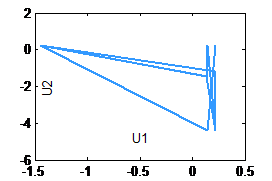}%
  \caption{Evolution of the strategies for the two players in the Kydland game under the GA: trajectories of decision variables across generations.}
  \label{fig:ga_strategies_kydland}
\end{figure}

\subsubsection{GA with Complete Information}

We then re-solve the same example under the assumption of complete information, parametrizing the strategies as a vector of open-loop variables. For different ranges of the variables (e.g., $[-1,1]$ vs.\ $[-5,5]$), the GA converges to the same Nash equilibrium, with different numbers of generations required. Increasing the crossover rate also accelerates convergence, as reported in \cite{WhitleyGATutorial,GenCheng1997}.

\subsubsection{PSO Results}

The PSO and hybrid PSO algorithms were also applied, both with and without mutation. Without mutation, PSO converges faster but is more sensitive to the initial conditions. With mutation and local search, the hybrid PSO converges reliably to the same Nash equilibrium across runs, with accuracy about $0.01$, similar to other hybrid PSO applications in engineering control and robotics \cite{Rezaee2017PID,Rezaee2017MPC,Ramezani2024Drones}.

In all cases, the velocities of the particles oscillate around the Nash equilibrium point and then converge to it.

\section{Conclusion}
\label{sec:conclusion}

We have presented two population-based evolutionary algorithms---a co-evolutionary genetic algorithm and a hybrid PSO method---for computing Nash equilibria in dynamic non-cooperative games. Compared with classical mathematical techniques \cite{BasarOlsder1999,Ozyildirim2000}, these methods:
\begin{itemize}
    \item Can be applied to games with nonlinear dynamics and non-quadratic costs,
    \item Are flexible with respect to the number of players and stages,
    \item Are less likely to be trapped in local Nash equilibria.
\end{itemize}

Our work is aligned with a broader trend in applying evolutionary and learning-based methods to high-dimensional, nonlinear problems across pattern recognition \cite{Hajati2006FaceLocalization,Pakazad2006FaceDetection,Hajati2010PoseInvariant,Hajati2017DynamicTexture,Hajati2017Surface,AbdoliHajati2014,Shojaiee2014Palmprint,CremersACCV2014}, biomedical and healthcare analytics \cite{Fiorini2019,Mahajan2024,Sadeghi2024COVID,Sopo2021DeFungi,Sadeghi2024ECG,Tavakolian2022FastCOVID,Tavakolian2022SoftwareImpacts,Wang2022SoftwareImpacts,Shahramian2013Leptin,Shahramian2013Troponin}, smart grids and energy forecasting \cite{Barzamini2012}, intelligent control and robotics \cite{RezaeeGolpayegani2012,Rezaee2017PID,Rezaee2017MPC,Rezaee2017Penetrometer,Rezaee2010FIR,Ramezani2024Drones}, communications and antenna design \cite{MohamadzadeRezaee2017Antenna,Taghvaee2014Metamaterial}, and large-scale distributed computing and cloud services \cite{Rezaee2014FuzzyCloud,Gavagsaz2018LoadBalancing,Rezaee2008GeneticSymbiosis}. 

Future work may consider more advanced co-evolutionary schemes, alternative neighborhood structures in PSO, and hybridization with other optimization methods---including reinforcement learning and model predictive control \cite{Rezaee2017MPC,Ramezani2024Drones}---to further improve convergence and robustness. In addition, integrating behavioral models from cognitive and educational studies \cite{Sarvghad2011ThinkingStyles} and leveraging large-scale networked platforms \cite{BarolliAINA2024,BarolliBWCCA2019,BarolliWAINA2019} may open new directions for multi-agent game-theoretic modeling in complex socio-technical systems.

\bibliographystyle{unsrt}

\end{document}